%% file: pbnclass2.tex
\newif\ifeusipstyle
\newif\ifdohybrid
\eusipstylefalse
\dohybridfalse

\ifeusipstyle
  \documentclass[a4paper]{article}
  \usepackage{spconf,amsmath,graphicx,cite}
\else
  \documentclass[conference]{IEEEtran} 

  \usepackage[dvips]{graphicx}
\fi

\usepackage{url}
\usepackage{amsfonts}

\title{The Projected Belief Network Classfier : both Generative and Discriminative.}

\ifeusipstyle
   \name{Paul M Baggenstoss}
   \address{Fraunhofer FKIE, Fraunhoferstr 20,
   \\ 53343 Wachtberg, Germany}
\else
   \author{\IEEEauthorblockN{Paul M. Baggenstoss}
   \IEEEauthorblockA{Fraunhofer FKIE,
   Fraunhoferstrasse 20\\
   53343 Wachtberg, Germany\\
   Email: p.m.baggenstoss@ieee.org}
   }
\fi

\begin{document}
\include{macros}
\newtheorem{identity}{Identity}
\newtheorem{hypothesis}{Hypothesis}
\newcommand{\mathtiny}[1]{\mbox{\tiny$#1$}}

\maketitle

\begin{abstract}
The projected belief network (PBN) is a layered generative network
with tractable likelihood function, and is based on a feed-forward
neural network (FF-NN). It can therefore share an embodiment with a
discriminative  classifier and can inherit the best qualities of both 
types of network.  In this paper, a convolutional PBN is constructed that is both 
fully discriminative and fully generative and is tested on spectrograms of spoken commands.
It is shown that the network displays excellent qualities from either
the discriminative or generative viewpoint.  Random data synthesis and
visible data reconstruction from low-dimensional hidden variables 
are shown, while classifier performance approaches that of a 
regularized discriminative network.
Combination with a conventional discriminative CNN is also demonstrated.
\end{abstract}

\section{Introduction}
\subsection{Background and Motivation}
Much has been published on the comparison of generative and discriminative classifiers.
The widespread view is that discriminative classifiers
generalize better when sufficient labeled training data is available \cite{Lasserre06}.
Despite their success,  it has been recognized that
discriminative methods have flaws, vividly demonstrated by
{\it adversarial sampling} \cite{MayerAdvSamp}, a technique in which
small, almost imperceptible changes to the input data cause
false classifications.  Because generative classifiers are based on a model of the
underlying data distribution, they are less succeptible to adversarial sampling
and can complement discriminative classifiers.
There are a large number of methods that seek to combine generative and discriminative
classifiers \cite{jaakkola98exploiting,raina03classification,fng01,Fujino05,Holub08,Bosch08,Lasserre06},
or to combine discriminative and generative training \cite{Lasserre06,Minka05,BishopGenDisc}.
The weakness of generative classifiers stems from the need to estimate the data distribution,
a very difficult task that is unecessary when just classifying between known
data classes \cite{Vapnik99}.  In modeling complex data generation processes found in real-world
data, traditional probability density function (PDF) estimators such as kernel mixtures and hidden Markov models
do not suffice.   Deep layered generative networks (DLGNs) can model complex generative
processes, but the data distribution, also called likelihood function (LF) is intractable,
complicating training and inference.
Reason: the hidden variables are jointly distributed with the input data and must be integrated out.
Such networks need to be trained using surrogate cost functions such as contrastive divergence 
to train restricted Boltzmann machines \cite{WellingHinton04,HintonDeep06}, and Kullback
Leibler divergence to train variational auto-encoder (VAE) \cite{pmlr-v32-rezende14},
or an adversarial discriminative network to train generative adversarial networks 
(GAN)  \cite{GoodfellowGAN2014}.  Using DLGNs as classifiers is problematic not just
because of the intractability of the LF, but because the performance of generative
models in general lags behind discriminative classifiers. 
Significant performance improvements could potentially be made by combining
DLGNs with deep discriminative networks. But, in 
order to see a benefit by combining, all classifiers need to have good performance. 
Therefore, having a DLGN classifier with comparable 
performance to deep discriminative network would be greatly desireable.
In summary, there is a need for a DLGN with tractable LF  
that can be combined with discriminative approaches.
The newly introduced layered generative network called projected belief network (PBN) 
stands out as a potentially better choice to achieve these goals.
The PBN is a DLGN, so can model complex generative processes,
but stands out from all other DLGNs.  The tractable LF allows
direct gradient training and enables detection of out-of-set samples 
(outliers that are outside of the set of training classes).
Because the PBN is based on a feed-forward neural network (FF-NN), it can share 
an embodiment with a discriminative classifier (i.e. it is a single network
that is both a complete generative model and a discriminative classifier),
so is a more direct way to introduce the
advantages of generative models into a discriminative classifier, or vice-versa.

\subsection{Main Idea}
\begin{figure}[h!]
  \begin{center}
    \includegraphics[width=3.5in]{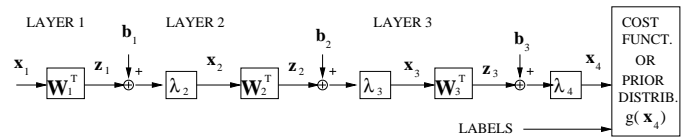}
  \caption{A feed-forward neural network (FF-NN).  This FF-NN can be a discriminative classifier
if $\lambda_4$ is the {\it softmax} function and the output
box is the cross-entropy cost function.  It can also be a generative model if viewed as  PBN
 and  the output box is the output prior distribution  $g(\bfx_4)$.}
  \label{asy0}
  \end{center}
\end{figure}

The PBN is based on a feed-forward neural network (FF-NN).
Figure \ref{asy0} shows a simple 3-layer FF-NN.
Each layer $l$ consists of a linear transformation (represented by matrix ${\bf W}_l$),
a bias $\bfb_l$ and an activation function $\lambda_{l+1}(\;)$.
The linear transformation can be fully-connected or convolutional,
but must have total output dimension equal to or lower than the input dimension.
This network can serve as a traditional classifier network if the output layer dimension is 
equal to the number of classes (and the final activation function  is {\it softmax}).
On the other hand, it can also be viewed as a projected belief network (PBN) \cite{BagPBN,BagEusipcoPBN}
whose properties are reviewed below.

\subsection{Mathematical Foundation}
Suppose we are given a fixed dimension-reducing transformation mapping
high-dimensional input data $\bfx\in \mathbb{R}^N$ to 
a lower-dimensional feature $\bfz\in \mathbb{R}^M$, $M<N$, denoted by $\bfz = T(\bfx)$.
If the probability density function (PDF) of the feature, denoted by $g(\bfz)$,
is estimated or specified, then we may ask the question ``given
$g(\bfz)$ and $T(\bfx)$, what is a good estimate of the PDF of $\bfx$?". 
The method of maximum entropy (MaxEnt) PDF projection \cite{Bag_info}
finds a unique PDF defined on $\mathbb{R}^N$ with highest
entropy among all PDFs consistent with $T(\bfx)$ and $g(\bfz)$. 
When PDF projection is applied layer-wise to a feed-forward neural network (identified with $T(\bfx)$) then 
a projected belief network (PBN) results \cite{BagPBN,BagEusipcoPBN}.
The LF for the network in Figure \ref{asy0} is given by (see \cite{BagEusipcoPBN}) 
\beq
\begin{array}{l}
p_p(\bfx_1; T, g) = 
\frac{1}{\ds \epsilon} \; \frac{\ds p_1(\bfx_1)}{\ds p_1(\bfz_1)} \;  |{\bf J}_{\bfz_1 \bfx_2}|  \\  
  \;\;\;\;\;\;\;\;\; \cdot \; \frac{\ds p_2(\bfx_2)}{\ds p_2(\bfz_2)} \; |{\bf J}_{\bfz_2 \bfx_3}| \;
	\frac{\ds p_3(\bfx_3)}{\ds p_3(\bfz_3)} \;  |{\bf J}_{\bfz_3 \bfx_4}| \; g(\bfx_4),
\end{array}
\label{cr1a}
\eeq
where $p_l(\bfx_l)$ is the assumed prior distribution for the input to layer $l$, 
 $p_l(\bfz_l)$ is the distribution of $\bfz_l$ under the assumption that
$\bfx_l$ is distributed according to $p_l(\bfx_l)$, 
$|{\bf J}_{\bfz_l \bfx_{l+1}}|$ is the determinant of the Jacobian (matrix of gradients)
of the invertible transformation mapping $\bfz_l\rightarrow \bfx_{l+1}$, and where $g(\bfx_{L+1})$ is the assumed prior for the output of a network.
The constant $\epsilon$ is the {\it sampling efficiency} discussed below.

\subsection{PBN layers and MaxEnt Priors}
The properties of PBN layer depend greatly on the assumed prior distribution $p_l(\bfx_l)$,
which is selected using the principle of Maximum Entropy (MaxEnt) and depends on the assumed input data 
range for the layer \cite{BagIcasspPBN}.
Consider a generic PBN layer with input dimension $N$.
There are three canonical input data ranges, {\it unlimited} denoted by $\mathbb{R}^N$,
{\it positive quadrant} where $0 < x_i$ denoted by $\mathbb{P}^N$,
and the {\it unit hypercube} where $0 < x_i < 1$ denoted by $\mathbb{P}^N$.
The MaxEnt priors for these data ranges are given in Table \ref{tab1v}.
For each data range and prior, there is a prescribed
input non-linearity (activation function) $\lambda(\;)$ which is also given in the table and should be applied at the output
of the previous layer.
The TG and TED activations resemble {\it softplus} and {\it sigmoid}, respectively \cite{BagIcasspPBN}.
Dimension-preserving layers are also possible ($M=N$) and are analyzed using the
determinant of the Jacobian matrix of the transformation.
\begin{table}
\begin{center}
 \begin{tabular}{|l|l|l|}
\hline
	 ${\cal X}$ &  MaxEnt Prior $p_0(\bfx)$  & $\lambda(\alpha)$ \\
 \hline
	 $\mathbb{R}^N$   & $\prod_{i=1}^N {\cal N}(x_i)$  (Gaussian) & $\alpha$  (Linear)\\
 \hline
	 $\mathbb{P}^N$   & $\prod_{i=1}^N 2 {\cal N}(x_i), \;\; 0<x_i $ (TG) & $\alpha + \frac{{\cal N}(\alpha)}{\Phi(\alpha)}$ (TG)\\
 \hline
	 $\mathbb{U}^N$   & $1, \;\; 0<x_i<1$  (Uniform) & $\frac{e^{\alpha}}{e^{\alpha} - 1}-\frac{1}{\alpha}$ (TED)  \\
 \hline
\end{tabular}
\end{center}
\caption{MaxEnt priors and activation functions as a function of input data range.
	TG=``Trunc. Gauss.". TED=``Trunc. Expon. Distr".
${\cal N}\left(x\right) \defined \frac{e^{-x^2/2}}{\sqrt{2\pi}}$ and $\Phi\left( x\right)  \defined \int_{-\infty}^x {\cal N}\left(x\right).$
}
\label{tab1v}
\end{table}

%
The sampling efficiency $\epsilon$ is less than 1 if the
output range of a layer is not identical to the assumed
input range of the next layer resulting in {\it subspace mismatch}.
However, $\epsilon$ is driven closer to 1 as the network trains
\cite{BagEusipcoPBN}, so can be ignored (i.e. assumed to be 1.0)  for all practical purposes.  
It is also possible to create PBNs with sampling efficiency identically equal to 1 by
using layers with Gaussian prior and 
dimension-preserving layers ($M=N$), both of which have no subspace mismatch.
The PBN is trained by maximizing the mean of the log of (\ref{cr1a})
using stochastic gradient ascent.  

\section{Technical Approach}
\subsection{Output Non-Linearity and Prior}
In order to create a PBN that is also a discriminative classifier,
a label-dependent output non-linearity and output prior are required.
A simple approach would be to apply a label-dependent level-shift
to the output variables, then assume a zero-mean standard normal output prior.
Specifically, one applies a level-shifting function $\lambda(z_i) = z_i-l_i$, $1\leq i \leq M$,
where $z_i$ is the network output (prior to activation function),
where $M$ is the number of data classes and the dimension of the network output, and 
${\bf l}=[l_1,l_2 \ldots l_M]$ is the label signal, a 
shifted one-hot encoding of the ground-truth label, with elements taking values of
$-L$ or $L$. Then, training with the simple Gaussian prior $g(\bfx_{L+1}) = {\cal N}(\bfx_{L+1}),$
where ${\cal N}(\bfx)=-\frac{M}{2}\log(2\pi)-\frac{1}{2} \bfx^\prime \bfx$ 
encourages the network output to agree with the label signal.
But, this approach makes little penalty for classification errors.  The function
$$
  \lambda(z_i) = z_i+C[\sigma(3 z_i)-.5]-l_i*(L+C/2)/L,
$$
where $\sigma(\;)$ is the sigmoid function and $C$ is a large constant,
produces a dynamic level shift that greatly increases if the
network output has the wrong sign (compared to the label).
When combined with the standard normal prior, it encourages 
Gaussian modes at -$L$ and +$L$, but imposes a very large penalty for class errors.
The degree of discriminative training can be varied by changing $C$.  

\subsection{MaxEnt Reconstruction and Synthesis}
\label{reconsec}
We now investigate a distinctly generative property of the
PBN : visible data reconstruction from hidden variables.
Input data can be randomly synthesized or
reconstructed from the output of any layer of the FF-NN.
Unlike other generative networks, the PBN is not an
explicit generative network, it operates implicitly 
by ``backing up" through a FF-NN.
In each layer,  the PBN selects a sample from the set
$${\cal M}(\bfz) = \{ \bfx : {\bf W}^\prime \bfz=\bfx\},$$
which is the set of samples $\bfx$ that ``could have" produced
$\bfz$.  A sample is selected from ${\cal M}(\bfz)$
with probability density proportional to the prior distribution $p_0(\bfx)$. 
When $p_0(\bfx)$ is the uniform
distribution, this is called uniform manifold sampling (UMS)  \cite{BagUMS}.
Sampling requires a type of Markov chain Monte-Carlo (MCMC) \cite{BagUMS}.
Deterministic data generation is also possible if instead
of randomly selecting a sample in ${\cal M}(\bfz)$, we
select the mean (the conditional mean given $\bfz$) ,
denoted by $\hat{\bfx}|\bfz = {\mathbb E}(\bfx|\bfz).$
This can be found in closed form for
a range of MaxEnt priors \cite{BagIcasspPBN,BagEusipcoPBN,BagUMS}
and  is given by $\hat{\bfx}|\bfz =  \lambda({\bf W} \hat{\bfh}),$
  where $\lambda(\;)$ is given in Table \ref{tab1v} and
$\hat{\bfh}$ is the solution of the equation 
\beq
{\bf W}^\prime \lambda\left({\bf W} \bfh \right)=\bfz.
\label{hsola}
\eeq
This solution is guaranteed to exist as long as $\bfx$ is in the support $p_0(\bfx)$
and is also the saddle-point for the saddle-point approximation to $p_0(\bfz)$  \cite{BagIcasspPBN}.
For the simplest case of Gaussian MaxEnt prior, the activation function is linear, $\lambda(\alpha)=\alpha$,
and the reconstruction is by least-squares, $\hat{\bfx}|\bfz = {\bf W} \left({\bf W}^\prime {\bf W}\right)^{-1}\bfz.$

Starting at any layer output, one can proceed in the backward direction
up the network, increasing the dimension, until the visible data
is reconstructed.  There are two possible reconstruction methods, (a) 
random sampling in ${\cal M}(\bfz)$ by MCMC, and (b) deterministically
selecting the conditional mean   $\hat{\bfx}|\bfz$.
When subspace mismatch occurs,
the reconstruction chain could fail. The rate of success is the
{\it sampling efficiency} discussed above, and is different for
stochastic and deterministic reconstruction.
Generally, a trained network has a deterministic sampling efficiency of 1
(failure is rare or non-existent).
Reconstructing from dimension-preserving layers involves just a matrix inversion,
so has a sampling efficiency of 1.
Layers with Gaussian input assumption also have a sampling efficiency of 1.
Deep networks can be constructed using these two layer types to obtain
deep PBNs with sampling efficiency of 1.

When only determintstic reconstruction is used, the result is a deterministic
PBN \cite{BagEusipcoPBN}, a type of auto-encoder where the reconstruction
network that is defined by the analyis network.

\subsection{PBN Properties}
The PBN differs significantly from other methods
of combining the roles of generative and discriminative networks because the
discriminative influence is added into the output prior
and does not disturb the ``purity" of the generative network.
There is no compromise between generative and
discriminative training or structure, they are both 
contained in one network and one cost function.

When reconstructing visible data from hidden variables, 
the synthesized data, when applied to the feed-forward
network, produces exactly the same hidden variables as were
created during the generation process. This property of hidden variable recovery
is unique to the PBN.  

During training, when the discriminative cost function is ``satisfied" (the training data is almost 
completely separated), then the generative cost dominates, so the network becomes the best possible
PBN that at the same time separates the data.  This can be seen as a generative regularization effect.

\section{Classification of Spectrograms of Words Commands}
\subsection{Data set}
The data was selected to be at the same time relevant, realistic,
and challenging.  We selected a subset of the Google speech commands data \cite{GoogleKW},
choosing three pairs of difficult to distinguish words: ``three, tree",
``no, go", and ``bird, bed", sampled at 16 kHz and segmented into 
into 48 ms Hanning-weighted windows shifted by 16 ms.  
We used log-MEL band energy features with 20 MEL-spaced
frequency bands and 45 time steps, representing a frequency span of 8 kHz and a time span of 0.72 seconds.
The input dimension was therefore $N=45\times 20=900.$
From each of the six classes, we selected 500 training samples, 150 validation samples, at random.
The remaining samples were used to test, averaging about 1500 per class or about a total of 10000
testing samples.

\subsection{Network}
A separate network was trained on each word pair.
The networks had $L=5$ layers.  The first layer was convolutional with
9 ($21\times 17$) convolutional kernels using ``same" border mode and
$5\times 4$ downsampling (not pooled, just down-sampled), thus producing
9 ($9\times 5$) output feature maps, or a total output dimension of 405.
The second layer was convolutional with
24 ($5\times 3$) convolutional kernels using ``same" border mode and
$2\times 2$ downsampling, thus producing
24 ($3\times 2$) output feature maps, or a total output dimension of 144.
The remaining two layers were fully-connected with 64, and 24 neurons.
The output layer had 2 neurons, matching the number of classes. 
Note that we sought to reduce the dimension in each layer by at least a factor of 2.
The layer output activation functions were linear, linear, TG, TG, and linear 
(See Table \ref{tab1v}).

\subsection{Classification Results}
For a classifier benchmark, we trained a conventional CNN classifier network
on each class pair.  Each network consisted of
seven layers, three convolutional and four dense layers.
The convolutional layers had kernel shapes of $(11\times 5)$, $(7\times 5)$, and $(3\times 3)$,
max-pooling of  $(5\times 2)$, $(3\times 2)$, and $(1\times 1)$,
with 64, 32, and 48 kernels, respectively.
Soft-plus activation and ``same" convolutional padding was used.
The dense layers had  256, 128, 32 , and 2 neurons.
Dropout and batch normalization were used with ADAM optimization.
Classification accuracy for the CNN is given in Table \ref{tab1}.

The PBN networks were first initialized with random weights and 
trained as a standard discriminative deep neural network (DNN)
with dropout and L-2 regularization.
No data augmentation (such as random shifting) was used.
Classification accuracy for the initial PBNs is given in Table \ref{tab1} as `PBN(DNN)".
\begin{table}[htb]
\begin{center}
 \begin{tabular}{|l|l|l|l|}
\hline
 & \multicolumn{1}{|c|}{"three-tree"} &  \multicolumn{1}{|c|}{"no-go"}  & \multicolumn{1}{|c|}{"bird-bed"} \\
 \hline
         CNN &    0.923 &  0.924  & 0.965 \\

 \hline
	 PBN(DNN) &  0.873 &  0.859 &  0.960 \\
 \hline
	 PBN &  0.886 &  0.863 &  0.960 \\
 \hline
	 PBN+CNN &  {\bf 0.925} &  {\bf .926} &  {\bf 0.971} \\
 \hline
\end{tabular}
\end{center}
	\caption{Classification accuracy for the three class pairs.}
	\label{tab1}
\end{table}
The initialized PBN were then trained 
as a PBN by maximizing the mean likelihood function (\ref{cr1a})
with output prior distribution parameter  $C=200$ and L2 regularization.
The classification results for the PBN on the three class pairs are shown
in Table \ref{tab1} where they can be compared with the initial
DNN-trained networks ``PBN(DNN)". Note that
the PBN has about the same accuracy as ``PBN(DNN)", with slightly higher accuracy for two class pairs.
This demonstrates that the PBN training does not seem to impact the classifcation
performance of a network, and may even help.
Training a classifier network as a PBN can be regarded as a form of regularization.

\subsection{Reconstruction Results}
\begin{figure}[h!]
  \begin{center}
    \includegraphics[width=3.5in,height=1.0in]{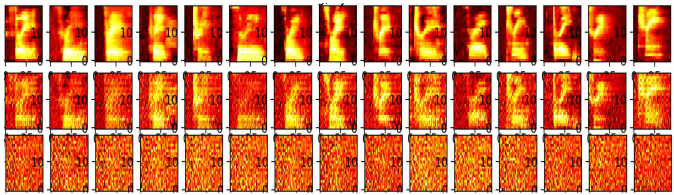}
  \caption{Samples of spoken word commands 
	``three" and ``tree" and reconstruction using the discriminatively-trained network.
	  From top: original samples, first-layer reconstructions, 
	  second-layer reconstructions.  }
  \label{dnnrecon}
  \end{center}
\end{figure}

It has been established that the PBN has lost little in terms of
classification performance wwhen compared to the
initial regularized discriminative networks. It will now be determined
what has been gained in terms of generative power.
The first thing that comes to mind is the reconstruction of visible data
from the hidden variables.  Using the method of Section \ref{reconsec},
we reconstructed data from the hidden variables
of the first and second layer of the initial PBN ``PBN(DNN)", with dimensions
405 and 144, respectively.  Results are shown in Figure \ref{dnnrecon}.
Little resemblance can be seen despite the high dimension
of the hidden variables.  This is how the network sees the data through the hidden variables.
The noisy images, when used as input data will produce exactly
the same hidden variables at the given layer as the
original input sample, a disturbing fact that vividly illustrates one of the
problems with discriminative networks.

\begin{figure}[h!]
  \begin{center}
    \includegraphics[width=3.5in]{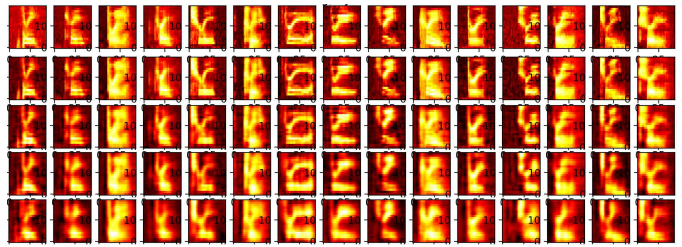}
  \caption{Samples of speech commands ``three, three" reconstructed using PBN. From top: original spctrograms, then the same reconsructed 
from output of first through fourth layers, with hidden variable dimensions of 405, 144, 64, and 24,
	  respectively.}
  \label{pbnrecon}
  \end{center}
\end{figure}

The reconstruction experiment was repeated for the trained PBN.
Results are shown in Figure \ref{pbnrecon}. 
This time, reconstruction was attempted from deep within the network.
The reconstructions had excellent quality, but gradually decreasing sharpness.  
Note that this network was not trained for lower reconstruction error, but instead to maximize (\ref{cr1a}).
The reconstruction power of the network comes as a side-effect and can be tapped 
into anywhere in the network.  

\subsection{Classifying between class pairs}
\begin{figure}[h!]
  \begin{center}
    \includegraphics[width=3.5in,height=1.0in]{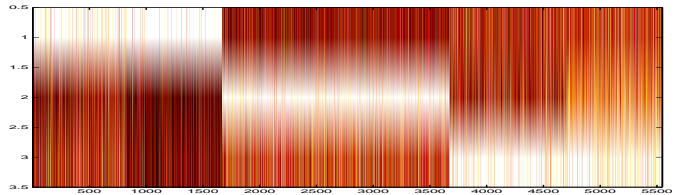}
	  \caption{Classifier statistic (negative log of 
	  mean square reconstruction error) for classifying between class pairs
	  based on reconstruction error.}
  \label{sixclass}
  \end{center}
\end{figure}

A second exercise in ``generative capability" is 
the classification between class pairs using models trained separately
on just one pair.  This demonstrates the ability to recognize out-of set events.
To classify between pairs, visible data reconstruction error was calculated
based on the 24-dimensional output of the fourth layer (not using the output layer),
and the model giving least error was chosen.
Figure \ref{sixclass} shows the classifier statistic (negative log of mean square reconstruction error).
The inter-pair classification accuracy was 87.9\%, which is good considering
the number of mal-formed events in the data base and
that the models were separately trained, without access to data
of the competing class pairs.  

\subsection{Combination with CNN}
\begin{figure}[h!]
  \begin{center}
    \includegraphics[width=3.5in,height=1.5in]{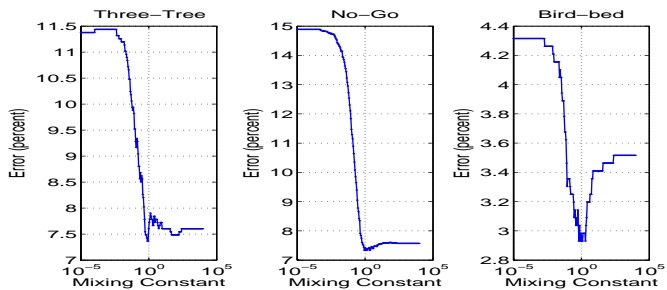}
  \caption{Classifier combination results for each class pair as a function
	  of linear combination weight. Performance at the
	  far right of each graph corresponds to CNN only
	  and far left to PBN.}
  \label{classcomb}
  \end{center}
\end{figure}

We have postulated above that having a generative classifier
with comparable performance to a discriminative one would
allow for performance gains when combining them.
To demonstrate this, the PBN was combined with the  CNN benchmark classifier
described above.  In Figure \ref{classcomb}, combined classifier error in percent is shown
as a function of additive combination weight for each class pair.
As might be expected, the class-pair in which the generative and
discriminative performance are the most similar (see Table \ref{tab1})
shows the most improvement.

\subsection{Random Synthesis}
\begin{figure}[h!]
  \begin{center}
    \includegraphics[width=3.5in]{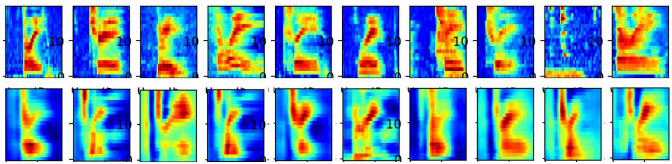}
  \caption{Top: ten training samples randomly selected from
	   ``three" and ``tree" spoken word commands. Bottom: randomly synthesized
	   data from trained PBN. There is no relationship to the
	   selected training samples on top.}
  \label{syn45}
  \end{center}
\end{figure}

As a final demonstration of generative power, we synthesized entirely random
events by starting with random data equal in dimension to the PBN
output layer, in this case dimension-2.
Data was synthesized at the point prior to the output
activation function using Gaussian random variables.
Results are shown in Figure \ref{syn45} for the class
pair ``three" and ``tree".
The synthetic samples appear realistic and are diverse,
showing variations in time shift, dilation, and other qualities.
This means that the PBN has indeed learned much about the
data generation process.

\subsection{Implementation and Applications }
The PBN was implemented in Python using Theano 
symbolic expression compiler \cite{Theano}.
The primary computational challenge is the solution
of a symmetric linear system with dimension $M\times M$,
where $M$ is the total output dimension of a layer.
This must be solved for each iteration in the solution
of (\ref{hsola}).  This was parallelized on the GPU, one processor
for sample in a mini-batch.  The computational time for an epoch was 1.1 seconds.
This was only about an order of magnitude slower than training the DNN.
All results were obtained using PBN Toolkit \footnote{http://class-specific.com/pbntk. A copy
of the data is also available at this link}.

\section{Conclusions }
In this paper, a projected belief network (PBN), which is a purely
generative layered network,  was trained as a
generative-discriminative classifier.  This was achieved using a label-dependent prior for the output features.
Since the PBN is based on a feed-forward neural network (FF-NN),  it can share
an embodiment with a discriminative deep neural network (DNN). 
Through the parameter $C$, the network can be trained with varying amount of discriminative influence.
When reconstructing visible data from the hidden variables, it was shown that the
the same netrork, trained discriminatively, had very poor ability to reconstruct, even from initial layers,
whereas whe the network was trained as a PBN, the reconstruction greatly improved.
The PBN classifier had comparable classification performance
to the discriminatively-trained network, yet provided generative power
from three standpoints: visible data reconstruction from hidden variables,
random data synthesis, and classification of out-of set samples.
It was also shown to improve upon a conventional CNN when
additively combined.

\bibliographystyle{ieeetr}
\bibliography{ppt}
\end{document}

\subsection{Background and Motivation }
Discriminative neural networks have dominated machine learning for decades.
The performance of generative networks lags behind 
because they need to model the generative process underlying the data, a much harder
task than discrimination \cite{Vapnik99}. Yet, interest in generative models persists
because a model of the underlying process is useful, as exemplified by variational
autoencoders (VAE) \cite{pmlr-v32-rezende14}, and generative adversarial network \cite{GoodfellowGAN2014}
(GAN) which have sparked considerable interest.
While the generative task is harder, given time and effort, 
generative models can perform as well as classifiers as
their discriminative counterparts. 
For example, when Hinton's deep belief network (DBN) 
was published, the DBN worked better than comparable fully-connected 
(non-convolutional) feed-forward networks \cite{HintonDeep06}.
While training algorithms have been developed for VAE and DBN,
the likelihood functions (LF) are not available in closed-form, so need to be approximated,
using stochastic variational methods in the case of VAE \cite{pmlr-v32-rezende14},
or Monte Carlo approximations in the case of DBN \cite{SalakhutdinovDBN}.
%
%
The projected belief network (PBN) is 
a new type of generative network with tractable 
LF that generates data layer-wise from hidden variables similar to a 
 deep latent Gaussian model (DLGM).  But, in contrast to other
generative models, the PBN  is related to a feed-forward neural 
network (FF-NN)  by a duality relationship \cite{BagPBN}.  
The dual FF-NN, which is here called dual analysis network (DAN), 
exactly recovers the hidden variables
of the PBNs data generation process.  
%
%
With tractable LF, the PBN has the potential to enable a new
class of generative models and algorithms.

\subsection{Main Idea}
The projected belief network (PBN) was previously introduced as a dual counterpart
to a feed-forward neural network (FF-NN) \cite{BagPBN}.
The PBN is derived from a FF-NN by asking the following question: {\it knowing
the FF-NN and the distribution of the output variables (features) of the FF-NN,
what is the  maximum entropy (MaxEnt) distribution of the visible 
data consistent with the given features distribution?}
The PBN is the generative network that implements this MaxEnt distribution \cite{BagPBN}.
Not surprisingly, the PBN uses the same network weights as the FF-NN
from which it is derived, and employs a special 
``activation" function that gives it its unique properties.
A deterministic version of the PBN is created if instead of
generating random data in each layer, the conditional
mean is propagated.  The deterministic PBN is the complementary
network to the DAN and combined with the DAN forms a new
type of auto-encoder.

\subsection{Paper Contributions}
The PBN has been previously introduced \cite{BagPBN}. Novel contributions of this paper include
(a) experimental results comparing PBN with 
other models as a function of data dimension,
(b) the detailed description of a multi-layer PBN,
(c) the treatment of the issue of sampling efficiency,
(d) the conceptual comparison of PBN with the VAE,
and (e) the description of a deterministic PBN
and its application as an auto-encoder,
and experiments showing significant improvements
over a conventional auto-encoder of the same structure.

\section{Projected Belief Networks (PBN)}
\subsection{PBN Exact Form}
Figure \ref{pbn_multi} illustrates a two-layer PBN in its exact, asymptotic,
and deterministic forms.  It can be easily extended to more layers.
\begin{figure}[h!]
  \begin{center}
    \includegraphics[width=3.5in]{pbn_multi_b.eps}
  \caption{A 2-layer PBN in three forms,
exact, asymptotic, and deterministic, and the
corresponding dual analysis network (DAN).}
  \label{pbn_multi}
  \end{center}
\end{figure}
Near the bottom of the figure is the dual analysis network (DAN), a conventional
feed-forward network employing an activation function
$\lambda_n(\;)$ in layer $n$.
Optionally, an energy statistic (ES), denoted by $e=t(\bfx)$ is extracted from the input
of each layer.
The figure illustrates both data generation by different forms of the
PBN (left to right) and feature extraction by the DAN (right to left).
Data generation originates by a feature generating distribution
$g(\bfz_2)$, then continues layer by layer.  In layer $n$ of the exact form of the PBN, (top), the
activation function and bias (if used) are inverted, and the
feature $\bfz_n$ is presented to the ``UMS" block in which a 
sample $\bfx$ is drawn randomly from the set ${\cal M}_n(\bfz_n,e_n)$ defined by
\beq
   {\cal M}_n(\bfz_n,e_n) = \{ \bfx : {\bf W}_n^\prime \bfx = \bfz_n, \; t_n(\bfx)=e_n, \;\;\; \bfx \in {\cal X}_n \},
   \label{manifze}
\eeq
where ${\cal X}_n$ is the input range of layer $n$ and $e_n=t_n(\bfx)$ is the optional ES.
The sample $\bfx$ must be drawn with uniform distribution,
so that no member of ${\cal M}_n(\bfz_n,e_n)$ is more likely to be drawn than any other.
The sampling procedure is therefore called uniform manifold sampling (UMS) \cite{BagUMS}.

By the definition of UMS, the DAN will exactly recover the variables $\bfz_2$, $\bfz_1$. 
When the PDF of $\bfz_2$ is known, denoted
by $g(\bfz_2)$, then the PBN generates samples the PDF:
\beq
p_p(\bfx_1; T, g) = \frac{1}{\epsilon} \; \frac{p(\bfx_1 ; H_{0,1})}{p(\bfz_1 ; H_{0,1})} \;  |{\bf J}_{\bfz_1 \bfx_2}| \; \frac{p(\bfx_2 ; H_{0,2})}{p(\bfz_2 ; H_{0,2})} 
\; g(\bfz_2),
\label{cr1a}
\eeq
where $\bfx_n$ is the input data to layer $n$ ($\bfx_1$ is the visible data), 
$T$ represents the DAN, $|{\bf J}_{\bfz_1 \bfx_2}|$ is the determinant of the
Jacobian of the 1:1 mapping from $\bfz_1$ to $\bfx_2$,
and $\epsilon$ is the sampling efficiency, to be explained below.

Notice the absence of integral signs in (\ref{cr1a}) - the distribution
does not require integrating out the hidden variables, as is necessary
in other layered generative models.  This is due to the fact that the
hidden variables of the DAN are deterministically derived from the
visible data, not jointly distributed.   Note also that in (\ref{cr1a})  
there appears a set of reference distributions, one for each layer.
The distribution $p(\bfx_n;H_{0,n})$ is the maximum entropy (MaxEnt) reference distribution
for layer $n$ and $p(\bfz_n;H_{0,n})$ is the corresponding feature distribution\footnote{$p(\bfz_n;H_{0,n})$ is the theoretical PDF
of the layer output when the layer input is distribued according to $p(\bfx_n;H_{0,n})$.}.
This reference distribution depends on ${\cal X}_n$, the data range of layer $n$ input,
which in turn depends on the activation function used in the previous layer - note
that the input (visible data) is assumed to have been created using $\lambda_1(\;)$.
We consider three data ranges: $\mathbb{R}^N$, $\mathbb{P}^N$,
and $\mathbb{U}^N$, where $N$ represents input data dimension of a generic layer,
$\mathbb{R}^N$ is the unlimited case, $\mathbb{P}^N$ is the positive quadrant ($0\leq x_i$),
and $\mathbb{U}^N$ is the unit hypercube ($0\leq x_i \leq 1$).
%
The MaxEnt reference distribution for each data range ${\cal X}$ is given in Table \ref{tab1v}.
The primary computational challenge in computing (\ref{cr1a})
is calculating the denominator terms $p(\bfz_n;H_{0,n})$.  More is provided in the 
references \cite{BagPDFProj,BagNutKay2000,Bag_info,BagUMS,BagPBN,BagEusipcoRBM}.
\begin{table}
\begin{center}
 \begin{tabular}{|l|l|l|l|l|}
\hline
${\cal X}$ &  $p(x;\alpha)$ & $\lambda(\alpha)$  & $t(\bfx)$ & $p(\bfx;H_0)$\\
 \hline
$\mathbb{R}^N$   &  ${e^{-(x-\alpha)^2/(2\sigma^2)} \over \sqrt{2 \pi \sigma^2} }$ (Gauss.) & $\alpha$  & $\sum_i x_i^2$  
& $\frac{e^{-t^2(\bfx)/2}}{(2\pi)^{-N/2}}$\\
 \hline
$\mathbb{P}^N$  &  $\alpha e^{-\alpha x}$  {\hspace{.37in}} (Expon.) & $1/\alpha$    & $\sum_i x_i$  & $e^{-t(\bfx)}$ \\
 \hline
$\mathbb{U}^N$   &   $\left(\frac{\alpha}{e^{\alpha} - 1}\right)  \; e^{\alpha x}$   {\hspace{.12in}} (TED) & $\frac{e^{\alpha}}{e^{\alpha} - 1}-\frac{1}{\alpha}$    & none & 1\\
 \hline
\end{tabular}
\end{center}
\caption{Generating distributions $p(x;\alpha)$, expected value of generating distributions $\lambda(\alpha)$,
energy statistics (ES) $t(\bfx)$,  and reference hypotheses $p(\bfx;H_0)$
for for data ranges $\mathbb{R}^N$, $\mathbb{P}^N$, and $\mathbb{U}^N$.
This table concerns a single layer and ${\bf x}$ is assumed to be the
visible data for the given layer layer with dimension $N$ and range ${\bf x}\in {\cal X}$.
}
\label{tab1v}
\end{table}


Depending on the data range (see Table \ref{tab1v}) an ES might need to be extracted from each layer input. We describe the ES for completeness,
but no ES is needed for $\mathbb{U}^N$, and for $\mathbb{P}^N$, the 
ES can be incorporated into matrix ${\bf W}_n$, eliminating the need for an explicit ES.  For more about the ES, please consult the references \cite{Bag_info,BagUMS}.

Optionally, a bias and activation function can be appended to the DAN (bottom of Figure \ref{pbn_multi}),
producing feature $\bfx_3$. In this case, the data generation process begins with the generating distribution
$g(\bfx_3)$, and the activation function and bias must be inverted.  Also, 
equation (\ref{cr1a}) must be modified by replacing $g(\bfz_2)$ with $|{\bf J}_{\bfz_2 \bfx_3}| \;g(\bfx_3).$

%

\subsection{PBN Asymptotic Form}
It has been shown that the UMS sampling process can be closely approximated by a
network layer resembling a sigmoid belief network \cite{BagUMS}.
To arrive at the asymptotic PBN (see Figure \ref{pbn_multi}), 
the UMS blocks are replaced by a nonlinear function $\bfh_n = \gamma_n^{-1}(\bfz_n)$,
 matrix multiplication $\balpha_n = {\bf W}_n \bfh_n$, then generation
from distributions $p_n(x; \alpha)$, which
are given in Table \ref{tab1v} as a function of ${\cal X}_n$.  The expected value of these distributions
(given $\alpha$) is denoted by $\lambda_n(\alpha)$, which
corresponds to the activation functions used in the DAN
at the output of layer $n-1$.  
Interestingly, for $\mathbb{U}^N$, $\lambda_n(\alpha)$
 is the mean of the truncated exponential distribution (TED),
which is similar to the sigmoid function \cite{BagUMS}.
Central to the theoretical analysis of a PBN layer
is the function $\gamma_n(\bfh_n)  =  {\bf W}_n^\prime \lambda( {\bf W}_n \bfh_n).$
To compute a layer of a PBN, this function needs to be inverted:
$\bfh_n=\gamma_n^{-1}(\bfz_n),$ which requires  
an iterative algorithm, but
might have no solution (See Section
\ref{sampeff}).

\subsection{The PBN for $\mathbb{R}^{N}$ and Relationship to VAE}
The VAE is currently a well-studied generative model \cite{Goodfellow2016,pmlr-v32-rezende14}.
The ``variational" aspect of VAE has to do with approximating and training the LF,
but the VAE is essentially an implementation of DLGM \cite{pmlr-v32-rezende14}.
Thus, both PBN and VAE are layered generative models. The main difference
is that the PBN is based on an explicit feed-forward analysis network (the DAN),
so the latent variables can be  deterministically computed from the
visible data. So, once a visible data
sample has been generated by the PBN, all the hidden variables can then be exactly
recovered by a single pass of the DAN.
The VAE on the other hand is a stochastic layered generative model,
so the latent variables of the VAE are jointly distributed
with the visible data. For this reason the LF of the VAE is only available 
as an integral over the hidden variables.
But, this distinction is moot because when looking at the asymptotic form of the PBN,
an approximation that is very good as has been demonstrated \cite{BagUMS},
we see that the PBN {\it behaves} like a traditional layered stochastic generative model.

A network layer of a DLGM is composed of an arbitrary
non-linear function followed by additive correlated noise \cite{pmlr-v32-rezende14}.
A network layer of an asymptotic PBN, on the other hand,  is composed of 
a non-linear function $\gamma_n^{-1}(\bfz)$, followed by multiplication by matrix 
${\bf W}_n$, then the generating distributions are applied
to produce the output variables.  Function $\gamma_n^{-1}(\bfz)$ and the generating distributions  
depend on the range of the layer output variable and are given in Table
\ref{tab1v}.  When $\bfx \in \mathbb{R}^{N}$, the generating distribution is
Gaussian, and is implemented by adding independent Gaussian noise\footnote{This can be
easily extended to correlated noise by introducing a matrix multiplification 
between the layers.}.  This produces a type of DLGM.
But, the Gaussian noise in an asymptotic PBN must be added
after a linear transformation, whereas for DLGM it is added after an arbitrary transformation.
It is not clear what this distinction means to the ultimate
PDF estimation capability, and can only be discovered by future experiments. 
Note also that for the DLGM, the activation function is taken to be
part of the ``arbitrary non-linear function" , whereas in the 
PBN, the activation function $\lambda_n(\;)$ is defined for the
dual DAN, which determines the function  $\gamma_n^{-1}(\bfz)$ used in the PBN.
In holding to the MaxEnt principle, 
for a given data range ${\cal X}$, the activation function 
$\lambda()$ is fixed, and therefore $\gamma_n^{-1}(\bfz)$ is fixed.
But, if one is willing to give up this MaxEnt distinction, there is 
flexibility in choosing $\lambda()$  so long
as it is invertible (for example use {\it softplus}, not {\it relu}).

In summary, both DLGM and PBN are layered
generative networks and it is not clear from the
above comparison which structure is better or more general.
It is clear, however, that the PBN under special conditions
(i.e. for ${\cal X}=\mathbb{R}^N$) approximates a type of DLGM and has a closed-form LF
 which is especially efficient to compute for this case (see \cite{Bag_info} Section IV.C, page 2821).
Future work is planned to compare DLGM and PBN in practice.

\subsection{PBN Deterministic Form}
The deterministic form of the PBN is obtained from the
asymptotic form by replacing $p_n(x;\alpha)$ by their expected values $\lambda_n(\alpha)$.
Interestingly, $\lambda_n(\alpha)$ cancels  $\lambda^{-1}_n(\alpha)$,
leaving $\gamma^{-1}_n(\;)$ as the only non-linearities, except at the visible layer.
This resulting PBN is a deterministic dual to the DAN, which exactly recovers the hidden values.
An arbitrary activation function $\lambda_n(\alpha)$ can be used as long
as $\gamma_n(\bfh_n)$ is defined using the same function. 
Note that $\lambda_n(\alpha)$ must be invertible, so
activations functions like {\it softplus} can be used, but not {\it relu}.

\subsection{Sampling Efficiency}
\label{sampeff}
The sampling efficiency $\epsilon$ is the fraction of times that
the PBN successfully creates a sample of visible data and
depends on the feature generating distribution
$g(\bfz)$ and whether exact (UMS) or deterministic generation is used.
A sampling failure occurs in a UMS block if the set
${\cal M}_m(\bfz_n,e_n)$ has no members, or in the asymptotic or
deterministic PBN if $\gamma^{-1}_{n}({\bf z}_n)$ has no solution.
When sampling fails, it is necessary to re-start
the process by drawing another feature value.
Sampling efficiency, either for UMS or
for deterministic PBN, can be driven towards 1.0 though
training, as will be demonstrated below.

\subsection{PBN Initialization and Training}
\label{jft}
In order to initialze the PBN so it has high sampling efficiency,
the weight matrices should be initialized by principal component analysis (PCA)
of the input data prior to the activation function \footnote{When data is already 
constrained to the range $[0,\; 1]$, as it is
in the MNIST corpus, it is useful to ``gaussianify" the data, mapping to $\mathbb{R}^N$ prior to PCA analysis (See Section \ref{ddesc}).}.
Scaling and bias are then used to provide good ``activation" of $\lambda_n(\;)$.
%
%
In this paper, two types of PBN training are used - deterministic auto-encoder training
and maximum likelihood (ML) training.
In auto-encoder training, the 
DAN is combined with the deterministic PBN to form an auto-encoder (a clockwise circular path at the 
bottom of Figure \ref{pbn_multi}).  Training is accomplished using back-propagation
to minimize total square reconstruction error.
Note that the parameters appear in both PBN and DAN, so the derivative has two terms.
It is critical to have high sampling efficiency for 
auto-encoder training.  In the experiments, $\epsilon$ 
approaches very nearly 1.0 after the first training epoch, 
even for testing data.


In ML training, the log of equation (\ref{cr1a}) is trained for highest average value
by gradient ascent.
%
%
%
We used a special ``uniform assumption" training in which 
the optional activation function (bottom of Figure \ref{pbn_multi}) is applied
to compress the data to the range [0,1],  
and the feature distribution $g(\bfx_3)$ is ignored.
Ignoring the feature distribution is tantamount
to assuming that $g(\bfx_3)=1$, the uniform distribution.
Interestingly, by training this way, a network is produced that,
in fact, produces feature data $\bfx_3$ that is independent uniformly distributed - the 
simplifying assumption becomes fulfilled.

\section{Classification Experiments}
We now compare PBN with a Gaussian mixture model (GMM) in a simple 
classification task.

\subsection{Reduced MNIST Data Description}
\label{ddesc}
For the following experiments, just three characters
``3", ``8", and ``9", of the MNIST handwritten data corpus were used.
Four pixel down-sampling rates were chosen: 1:1, 2:1, 3:1, and 4:1,
resulting in 
data dimensions of 784, 196, 100, and 49.
Since MNIST pixel data is coarsely quantized in the range [0,1],
a dither was applied to the pixel values\footnote{For pixel values
above 0.5, a small exponential-distributed random value was subtracted,
but for pixel values below 0.5, a similar  random value was 
added.}.  To create data in $\mathbb{R}^N$, the inverse sigmoid function was
then applied in order to create ``gaussianified" data with most pixel values in the range -10 and 10.  

\subsection{The 1-layer PBN}
We revisit the 1-layer PBN, which was previously introduced
\cite{BagPBN}. The results of 1-layer PBN experiments 
are relevant to determine if the PBN should be exended to a second layer.
In a multi-layer PBN, a given layer acts as a PDF
model for the features of the up-stream layer. So,
it seems that there is no advantage to adding a layer to a PBN
if a GMM works better than the added layer.
The idea, then is to test a 1-layer PBN against
a GMM as a function of dimension.
This experiment is data-set dependent, so the results
here apply only to MNIST.
%
As a performance benchmark, the GMM was 
applied to the ``gaussianified" data in $\mathbb{R}^N$,
using both diagonal (GMM-D), and full (GMM-F) covariance matrices
\footnote{To avoid singularities, the diagonal elements of the covariance
matrices were multiplied by the factor $(1+\delta)$, where 
$\delta=$ 0.3, 0.3, 0.5, and 0.6 for dimensions 49, 100, 192, and 784, respectively.}.
A separate 1-layer PBN was initialized using PCA,
then trained for each data class 
to maximize the mean log-likelihood using gradient ascent
with ``ADAM" optimization and  L2 regularization using ``uniform assumption" 
training (Section \ref{jft}).  After training, the final activation function was removed,
then $g(\bfz_1)$ was modeled as a GMM.
%
For $N=49, 100, 196, 784$, the number of hidden units
(columns of matrix ${\bf W}$) were 12, 16, 30, and 34, respectively.

Results of the experiment are shown in Figure \ref{pbn_N}.
The PCA-initialized PBN, with no further training
are reported as ``PBN-P", and with training as ``PBN-G".
When comparing ``PBN-P" with ``PBN-G", we can conclude
that ML training greatly improves a PBN.  This means that 
the PDF model offered by a 1-layer PBN is more than 
a just a re-packaged type of Gaussian model or PCA.
The next observation is that the PBN performs better than GMM-F above $N=100$.
%
%
\begin{figure}[h]
  \begin{center}
    \includegraphics[width=3.4in,height=2.4in]{pbn_N.eps}
  \caption{Model comparison as a function of data dimension.} 
  \label{pbn_N}
  \end{center}
\end{figure}
%
%
Both  GMM-F and PBN can model pixel correlation, GMM-F explicitly
using the covariance matrices, and PBN implicitly 
by decorrelating the features, as was noted at the end of Section \ref{jft}.
But, PBN requires $MN$ parameters, versus  the $MN^2$ parameters required for the GMM.
This may explain the advantage of PBN above $N=100$.  
The average sampling efficiency for PBN-G 
was 0.72, 0.85, 0.77, and 0.52 for $N=49,100,192,784$,
respectively.  The worst case change in per-pixel log-likelihood,
is 0.007, so sampling efficiency in Figure \ref{pbn_N} can be essentially
ignored.

\subsection{Multi-layer PBN}
The 1-layer PBNs for $N=196$ and $784$
were extended to a second layer with $16$ and $18$ hidden units, respectively.
The 2-layer PBNs were then trained with an assumption of uniform distribution for $g(\bfx_3)$, then the final activation function was removed and
GMM was used to model the final feature PDF $g(\bfz_2)$.  
Sampling efficiencies were 0.55 and 0.70, respectively,
also negligible.  Performance is shown in Figure \ref{pbn_N} as ``PBG-2-G"
and shows worse performance with respect to 1-layer PBN-G.
This could have been predicted based on Figure
\ref{pbn_N} because the feature dimension is much less than $100$.
Extending the PBN to a second layer would only be effective if the
first layer feature dimension is much larger. 
%

\section{Auto-Encoder (A-E) Experiments}
In the next experiment, a multi-layer
deterministic PBN together with the DAN
are used as an A-E and compared with a standard A-E network 
of the same structure. 
The full $28\times 28$ ($N=784$) data was used.
Separate A-Es were trained on each data 
class to minimize total square error by back-propagation.
ADAM optimization and L2 regularization was used for both network types.
TED (T), sigmoid (S) and softplus (P) activation functions were tried.
The average squared error was measured for testing and training data
and is listed in Table \ref{tab1a}. Although the 
conventional A-E attained a lower squared error
on the training data, it fared much worse on the test data. In contrast,
the PBN had similar squared error on both sets, significantly
out-performing the standard A-E - which
can probably be attributed to (a) that fact that the 
PBN uses the same weights for reconstruction and analysis, and thereby
implements the same task with half the parameters, and (b)
the reconstruction (PBN) is the perfect complement to the 
analysis network (DAN).
Using L2-regularization for conventional A-E did not change this.
The A-E performance for TED and sigmoid was similar, but training took longer for TED.
Sampling efficiency for PBN was 100 percent (no samples that failed reconstruction)
for training, and about 99.9\% (typically 1 sample or less failed) on the
test data.  
\begin{table}
\begin{center}
 \begin{tabular}{|l|l|l|l|l|l|l|}
\hline
Nodes & Act & Type & E-Train & E-Test & Class  \\
\hline
\hline
32-12 & T & A-E  & 7.40 & 10.39 & 1.94\% \\
\hline
32-12  & S & A-E  & 6.73 & 10.79 & 2.97\% \\
\hline
32-12 & T & {\bf PBN}  & 8.63 & {\bf 9.04} & {\bf 1.27\%}  \\
\hline
\hline
36-16 & T & A-E  & 5.84 & 8.21 & 2.57\% \\
\hline
36-16  & S & A-E  & 5.26 & 8.26 & 2.51\% \\
\hline
36-16 & T & {\bf PBN}  & 6.96 & {\bf 7.40} & {\bf 1.70\%}  \\
\hline
\hline
32-16-9 & P & A-E  & 8.27 & 15.3 & 4.4\%  \\
\hline
32-16-9 & P & {\bf PBN}  & 9.95 & {\bf 11.25} & {\bf 0.90\%}  \\
\hline
\end{tabular}
\end{center}
\caption{Total square error for auto-encoder task. Activation functions
(Act) are TED (T), sigmoid (S) and softplus (P)}.
\label{tab1a}
\end{table}
The good generalization of the PBN A-E suggests
using it as a classifier based on minimum reconstruction
error, which we tried.  The results are shown in Table \ref{tab1a} 
in column ``Class".  PBN performed significantly better than A-E,
attaining a very respectable 0.9\%, which handily out-performs 
the standard PBNs in Figure \ref{pbn_N} (denoted by ``PBN A-E").

The deterministic PBN is also useful to generate entirely synthetic data,
In Figure \ref{aenc_syn_pbn}, examples were generated by training a 
GMM on the features (i.e. output of the DAN), then 
passing synthetic features through the PBN. 
The configuration ``32-16-9" with softplus activation was used.  The synthetic samples are sorted in order of decreasing likelihood
(starting from top left), demonstrating the a benefit of 
a tractable likelihood function.
The quality of these samples suggests using the deterministic PBN
in a generative adversarial network (GAN)  - but differing from
a standard GAN in the posession of a tractable LF.
\begin{figure}[h]
  \begin{center}
    \includegraphics[width=3.5in]{aenc_syn_pbn.eps}
  \caption{Data synthesized from determinisic PBN and sorted in order of decreasing
likelihood value.}
  \label{aenc_syn_pbn}
  \end{center}
\end{figure}

\ifdohybrid
\section{Hybrid PBN classifier}
The goal in this experiment is to combine the
results of the last 2 sections by forming a hybrid PBN classifier 
from the 1-layer PBN classifier and the PBN auto-encoder.

For the PBN portion of the hybrid, we used an annealed kernel mixture.
The idea of a PBN kernel mixture is that the features extracted by
a PBN trained on one class might have useful information
regarding another class - especially for poorly formed 
handwritten characters.
The class-specific feature 
mixture (CSFM) \cite{BagIWCCSP,BagAESModelMix,BagUMS}  
approximates the PDF of one class using a mixture of 
all the PBNs, increasing the information
available without increasing the feature dimension.
Annealing improves the linear mixing of the kernels
\cite{BagAESModelMix,BagUMS}.  
We form an annealed kernel mixture of the PBN PDFs (\ref{cr1a}) 
as follows:
\beq
f_m(\bfx; a) = \left(
\sum_{l=1}^c w_{l,m} \; p_p(\bfx; T_l,\smallmath{\hat{p}(\bfz_l|H_m)})^{1/a}
\right)^a,
\label{csfma}
\eeq
where $c$ is the number of classes ($c=3$ here), 
$\bfz_l = T_l(\bfx)$ represents the DAN trained on class $l$,
$\hat{p}(\bfz_l|H_m)$ is the feature PDF estimate for feature $\bfz_l$ and data class $m$,
and $a$ is a heurisic annealing parameter.
The weights are estimated using training data using,
$$\hat{w}_{l,m} = \frac{ \sum_i \; p_p(\bfx_i; T_l,\smallmath{\hat{p}(\bfz_l|H_m)})^{1/a}}
{\sum_i \; \sum_k \; p_p(\bfx_i; T_k,\smallmath{\hat{p}(\bfz_k|H_m)})^{1/a}}.$$
For the deterministic PBN auto-encoder portion of the hybrid, 
we used $h_m(\bfx;b) = {e^{-\|\bfx-\hat{\bfx}_m\|^2/b} \over
\sum_l e^{-\|\bfx-\hat{\bfx}_l\|^2/b}},$
where $\|\bfx-\hat{\bfx}_l\|^2$ is the square auto-encoding error
using auto-encoder trained on class $l$.
The complete hybrid classifier distribution is given by
$p(\bfx|H_m; a,b)=\frac{f_m(\bfx; a) \; h_m(\bfx;b)}{K_m(a,b)}$, where
$K_m(a,b)$ is the normalization constant 
that can be estimated using Monte Carlo integration (MCI)
with the un-annealed mixture $f_m(\bfx; a=1)$ 
acting as proposal distribution\footnote{ 
As a motivation for PBN, we noted that having a
tractable LF avoids the need for MCI, yet here we are using MCI. 
This is not a contradiction.  As dimension increases, the proposal distribution needs 
to be increasingly well matched to the function to be integrated.
To normalize a high-dimensional distribution outright,
for example using GMM as proposal distribution would fail.
But, MCI is useful to normalize 
high-dimensional distributions with tractable LF 
that have been slightly modified,  where the un-modified
distribution acts as proposal distribution.
} \cite{BagUMS}.

Prior to estimating $K_m(a,b)$, classification error was optimized over $a$ (Figure \ref{comb_b} left), without the term  $h_m(\bfx;b)$.
Then, using the value $a=1500$, optimized over $b$
(Figure \ref{comb_b} right), with a resulting minimum error of 0.77\%,
entered in Figure \ref{pbn_N} (left) as ``PBN-H".
With these values of $a$ and $b$, the normalization constant
$K_m(a,b)$ was estimated using Monte Carlo integration
and the normalized LF entered in Figure \ref{pbn_N} (right).
\begin{figure}[h]
  \begin{center}
    \includegraphics[height=1.9in,width=1.1in]{comb_a.eps}
    \includegraphics[height=1.9in,width=1.1in]{comb_b.eps}
    \includegraphics[height=1.9in,width=1.1in]{comb_b_10.eps}
  \caption{Classification error on reduced MNIST
as a function of  $a$ (left) and  $b$ (center), and on full MNIST
as a function of $b$ (right).}
  \label{comb_b}
  \end{center}
\end{figure}

As a final experiment, the hybrid classifier was tried on the full 10-character MNIST data set
to verify the above results and so that it could be compared with published work.
The classification error as a function of $b$ is shown in Figure \ref{comb_b}, right side, and attains a minumum error of 1.25\%
at the same value of $b$, which is comparable to state of the art
fully-connected (non-convolutional) discriminative classifiers not employing pre-processing,
image distortions or deskewing \cite{MNISTResults}.
\fi

\section{Conclusions}
In this paper, a multi-layer PBN has been described,
in its standard, asymptotic, and deterministic forms.
Experiments comparing a 1-layer PBN with a GMM
on a reduced subset of MNIST show that PBN out-performs GMM 
only above a dimension of about 100, which would
suggest using a 2-layer PBN when the output dimension of the first layer is large.
This paper also described a deterministic multi-layer PBN for the
first time and it has been experimentally found to be superior to a standard auto-encoder
when generalizing to test data both in terms of
reconstruction error and classifier performance.
\bibliographystyle{ieeetr}
\bibliography{ppt}
\end{document}

%% file: macros.tex
\newcommand{\defined}{\stackrel{\mbox{\tiny$\Delta$}}{=}}
\newtheorem{example}{Example}
\newtheorem{conclusion}{Conclusion}
\newtheorem{assumption}{Assumption}
\newtheorem{definition}{Definition}
\newtheorem{problem}{Problem}
\newcommand{\erf}{{\rm erf}}

\newcommand{\sst}{\scriptstyle }
\newcommand{\xparen}{\mbox{\small$(\bfx)$}}
\newcommand{\hojz}{H_{0j}\mbox{\small$(\bfz)$}}
\newcommand{\Hozj}{H_{0,j}\mbox{\small$(\bfz_j)$}}
\newcommand{\smallmath}[1]{{\scriptstyle #1}}
\newcommand{\Hoz}[1]{H_0\mbox{\small$(#1)$}}
\newcommand{\Hozp}[1]{H_0^\prime\mbox{\small$(#1)$}}
\newcommand{\Hozpp}[1]{H_0^{\prime\prime}\mbox{\small$(#1)$}}
\newcommand{\hoz}{\Hoz{\bfz}}
\newcommand{\hooz}{\Hozp{\bfz}}
\newcommand{\hoooz}{\Hozpp{\bfz}}
\newcommand{\smJ}{{\scriptscriptstyle \! J}}
\newcommand{\smK}{{\scriptscriptstyle \! K}}

\newcommand{\erfc}{{\rm erfc}}
\newcommand{\bitem}{\begin{itemize}}
\newcommand{\dsum}{{ \displaystyle \sum}}
\newcommand{\eitem}{\end{itemize}}
\newcommand{\benum}{\begin{enumerate}}
\newcommand{\eenum}{\end{enumerate}}
\newcommand{\bdm}{\begin{displaymath}}
\newcommand{\bfzro}{{\underline{\bf 0}}}
\newcommand{\bfone}{{\underline{\bf 1}}}
\newcommand{\edm}{\end{displaymath}}
\newcommand{\beq}{\begin{equation}}
\newcommand{\bea}{\begin{eqnarray}}
\newcommand{\eea}{\end{eqnarray}}
\newcommand{\cali}{ {\cal \bf I}}
\newcommand{\caln}{ {\cal \bf N}}
\newcommand{\barray}{\begin{displaymath} \begin{array}{rcl}}
\newcommand{\earray}{\end{array}\end{displaymath}}
\newcommand{\eeq}{\end{equation}}
\newcommand{\btheta}{\mbox{\boldmath $\theta$}}
\newcommand{\bTheta}{\mbox{\boldmath $\Theta$}}
\newcommand{\blam}{\mbox{\boldmath $\Lambda$}}
\newcommand{\beps}{\mbox{\boldmath $\epsilon$}}
\newcommand{\bdelta}{\mbox{\boldmath $\delta$}}
\newcommand{\bgamma}{\mbox{\boldmath $\gamma$}}
\newcommand{\balpha}{\mbox{\boldmath $\alpha$}}
\newcommand{\bbeta}{\mbox{\boldmath $\beta$}}
\newcommand{\balphascript}{\mbox{\boldmath ${\scriptstyle \alpha}$}}
\newcommand{\bbetascript}{\mbox{\boldmath ${\scriptstyle \beta}$}}
\newcommand{\bLambda}{\mbox{\boldmath $\Lambda$}}
\newcommand{\bDelta}{\mbox{\boldmath $\Delta$}}
\newcommand{\bomega}{\mbox{\boldmath $\omega$}}
\newcommand{\bOmega}{\mbox{\boldmath $\Omega$}}
\newcommand{\blambda}{\mbox{\boldmath $\lambda$}}
\newcommand{\bphi}{\mbox{\boldmath $\phi$}}
\newcommand{\bpi}{\mbox{\boldmath $\pi$}}
\newcommand{\bnu}{\mbox{\boldmath $\nu$}}
\newcommand{\brho}{\mbox{\boldmath $\rho$}}
\newcommand{\bmu}{\mbox{\boldmath $\mu$}}
\newcommand{\sigi}{\mbox{\boldmath $\Sigma$}_i}
\newcommand{\bfu}{{\bf u}}
\newcommand{\bfx}{{\bf x}}
\newcommand{\bfb}{{\bf b}}
\newcommand{\bfk}{{\bf k}}
\newcommand{\bfc}{{\bf c}}
\newcommand{\bfv}{{\bf v}}
\newcommand{\bfn}{{\bf n}}
\newcommand{\bfK}{{\bf K}}
\newcommand{\bfh}{{\bf h}}
\newcommand{\bff}{{\bf f}}
\newcommand{\bfg}{{\bf g}}
\newcommand{\bfe}{{\bf e}}
\newcommand{\bfr}{{\bf r}}
\newcommand{\bfw}{{\bf w}}
\newcommand{\calX}{{\cal X}}
\newcommand{\calZ}{{\cal Z}}
\newcommand{\bb}{{\bf b}}
\newcommand{\bfy}{{\bf y}}
\newcommand{\bfz}{{\bf z}}
\newcommand{\bfs}{{\bf s}}
\newcommand{\bfa}{{\bf a}}
\newcommand{\bfA}{{\bf A}}
\newcommand{\bfB}{{\bf B}}
\newcommand{\bfV}{{\bf V}}
\newcommand{\bfZ}{{\bf Z}}
\newcommand{\bfH}{{\bf H}}
\newcommand{\bfX}{{\bf X}}
\newcommand{\bfR}{{\bf R}}
\newcommand{\bfF}{{\bf F}}
\newcommand{\bfS}{{\bf S}}
\newcommand{\bfC}{{\bf C}}
\newcommand{\bfI}{{\bf I}}
\newcommand{\bfO}{{\bf O}}
\newcommand{\bfU}{{\bf U}}
\newcommand{\bfD}{{\bf D}}
\newcommand{\bfY}{{\bf Y}}
\newcommand{\bSig}{{\bf \Sigma}}
\newcommand{\test}{\stackrel{<}{>}}
\newcommand{\zmk}{{\bf Z}_{m,k}}
\newcommand{\zlk}{{\bf Z}_{l,k}}
\newcommand{\zm}{{\bf Z}_{m}}
\newcommand{\ssq}{\sigma^{2}}
\newcommand{\dint}{{\displaystyle \int}}
\newcommand{\ds}{\displaystyle }
\newtheorem{theorem}{Theorem}
\newcommand{\postscript}[2]{ \begin{center}
    \includegraphics*[width=3.5in,height=#1]{#2.eps}
    \end{center} }